\PassOptionsToPackage{pdfpagelabels=false}{hyperref} 
\documentclass{IOS-Book-Article}
\usepackage{mathptmx}
\DeclareMathAlphabet{\mathcal}{OMS}{cmsy}{m}{n}
\usepackage{soul}\setuldepth{article}
\def\hb{\hbox to 11.5 cm{}}
\usepackage{graphicx}
\usepackage{amssymb}
\usepackage{xcolor}
\usepackage{lineno}
\newtheorem{thm}{Theorem}
\newtheorem{definition}[thm]{Definition}
\begin{document}
\pagestyle{headings}
\def\thepage{}
\begin{frontmatter}
\title{On a Generalized Framework for Time-Aware Knowledge Graphs}
\runningtitle{On A Generalized Framework for Time-Aware Knowledge Graphs}
\author[1]{\fnms{Franz} \snm{Krause}%
\thanks{Corresponding Author. E-mail: franz.krause@uni-mannheim.de}
},
\author[1]{\fnms{Tobias} \snm{Weller}}
\and
\author[1]{\fnms{Heiko} \snm{Paulheim}}
\runningauthor{F. Krause et al.}
\address[1]{Data and Web Science Group, University of Mannheim, Germany}
\begin{abstract}
Knowledge graphs have emerged as an effective tool for managing and standardizing semistructured domain knowledge in a human- and machine-interpretable way. In terms of graph-based domain applications, such as embeddings and graph neural networks, current research is increasingly taking into account the time-related evolution of the information encoded within a graph. Algorithms and models for stationary and static knowledge graphs are extended to make them accessible for time-aware domains, where time-awareness can be interpreted in different ways. In particular, a distinction needs to be made between the validity period and the traceability of facts as objectives of time-related knowledge graph extensions. In this context, terms and definitions such as \textit{dynamic} and \textit{temporal} are often used inconsistently or interchangeably in the literature. Therefore, with this paper we aim to provide a short but well-defined overview of time-aware knowledge graph extensions and thus faciliate future research in this field as well.
\end{abstract}
\end{frontmatter}
\thispagestyle{empty}
\section{Introduction}
Knowledge graphs (KGs) and their integration into domain-specific use cases represent a topic that has been gaining popularity in recent research. Their inherent information is usually encoded in the form of triples $(h,r,t)$ $\widehat{=}$ $(head, relation, tail)$ where a node $h$ has the relation $r$ to another node or attributive literal $t$. KGs are used to improve the performance in areas like question answering~\cite{QA} and recommendation~\cite{recommendation} regarding various domains, e.g., industrial manufacturing~\cite{kg_for_ind4.0} and biomedicine~\cite{biomedicine}. Furthermore, extension approaches exist which aim at enriching triples with additional metadata~\cite{rdf-star,context-aware,triedf}, such as annotations or timestamps. However, discrepancies regarding terminology can be found in the literature. For example, the term knowledge graph is often used interchangeably, although enrichment by metadata usually cannot be assumed without loss of generality. 
Furthermore, especially in the context of time-aware knowledge graph extensions, frequently used terms such as \textit{dynamic}, \textit{temporal}, and \textit{static} are applied inconsistently.

Therefore, in this work, the distinction between standard knowledge graphs by means of stationary sets of triples~\cite{towards_kg,kg_survey} and time-aware KG extensions is elaborated. Reminiscent, mutable and incremental knowledge graphs are introduced as special cases of dynamic and temporal KGs. These definitions should ultimately serve to standardize time-aware KG extensions and thus facilitate future research in this field as well.
\section{Related Work}
\label{relatedwork}
State of the art knowledge graphs as sets of triples with entries $(h,r,t)$ already provide a limited possibility of expressing time-awareness by assigning additional time-related annotations to nodes~\cite{kg_survey}. For example, the information that the European Union (EU) was founded in $1951$ can be encoded by $(EU, founded, 1951)$ where $1951$ is an attributive timestamp literal, in this case the corresponding year. However, time-related node annotations are not sufficient to encode the information that the United Kingdom (UK) joined the EU in $1973$ since the year $1973$ refers to the edge $(UK,member,EU)$ and not to a single node. To provide the encoded triples with further information, general approaches like RDF*~\cite{rdf-star} already exist which assign additional metadata to the edges. In the following, we restrict ourselves to time-related metadata in order to standardize time-aware KG extensions. In fact, there are already numerous works dealing with this problem, but most of them consider successive applications such as KG embeddings~\cite{context-aware,temp_emb} or graph completion~\cite{temp_completion_1,temp_completion}. However, although knowledge graphs with additional time-related metadata within the edges are always considered, there is no generalized definition for this kind of encoding. As most of these approaches refer to such graphs as temporal KGs, we adopt this notion as well. In particular, temporal extensions represent a local form of time-awareness as timestamps are added to each edge individually.

Additionally, several works exist which consider entire knowledge graphs as being non-stationary, i.e. dynamic, which is to be interpreted as global time-awareness. Similarly to temporal KGs, these apply methods for standard knowledge graphs to dynamic KGs to make them accessible for areas such as KG embeddings~\cite{puTransE,DKGE} and KG completion~\cite{dyn_compl}. However, these approaches do not attempt to adapt the original methods to enriched knowledge graphs. Rather, they try to make previous models and results reusable in an efficient way so that, for example, full retraining of an embedding model is not required after the information encoded within the knowledge graph is updated.

To the best of our knowledge, no general definitions of temporal or dynamic KGs exist yet. Usually, respective assumptions are similar but not identical. Furthermore, there are several works where these terms are used interchangeably, inversely, or not at all. Finally, there is no well-defined approach for combining local and global time-awareness.
\section{Preliminaries}
In this work, the term knowledge graph is used as a generalized notion for approaches that manage semistructured data based on formal conceptualizations, e.g., ontologies, as well as collections of instantiation rules indicating the validity of a graph's topology, i.e., of its inherent edges. However, this generalization is based on the most common implementation form of a knowledge graph as a set of triples, which will be referred to as a standard KG. Furthermore, ontologies which conceptualize a domain by means of triples $(h,r,t)$ are referred to as static ontologies as they allow no further extensions of the triple structure such that facts are to be regarded as final and static. Moreover, a KG may be interpreted in both stationary and dynamic ways, i.e., we allow the consideration of time-related graph evolutions with respect to a set of timestamps $\mathcal{T}$. We assume $\mathcal{T}$ to be of a strict order, so that for $\tau, \tau' \in \mathcal{T}$ with $\tau \neq \tau'$ either $\tau < \tau'$ or $\tau' < \tau$ follows. Therefore, for $\tau < \tau'$ the timestamp $\tau$ occurred before $\tau'$. In this context, we also define the closure $\overline{\mathcal{T}} := \mathcal{T} \cup\{ -\infty, \infty \}$ such that $-\infty < \tau < \infty$ holds for all $\tau \in \mathcal{T}$.
\begin{figure}[t!]
\centering
\includegraphics[width=0.8\linewidth]{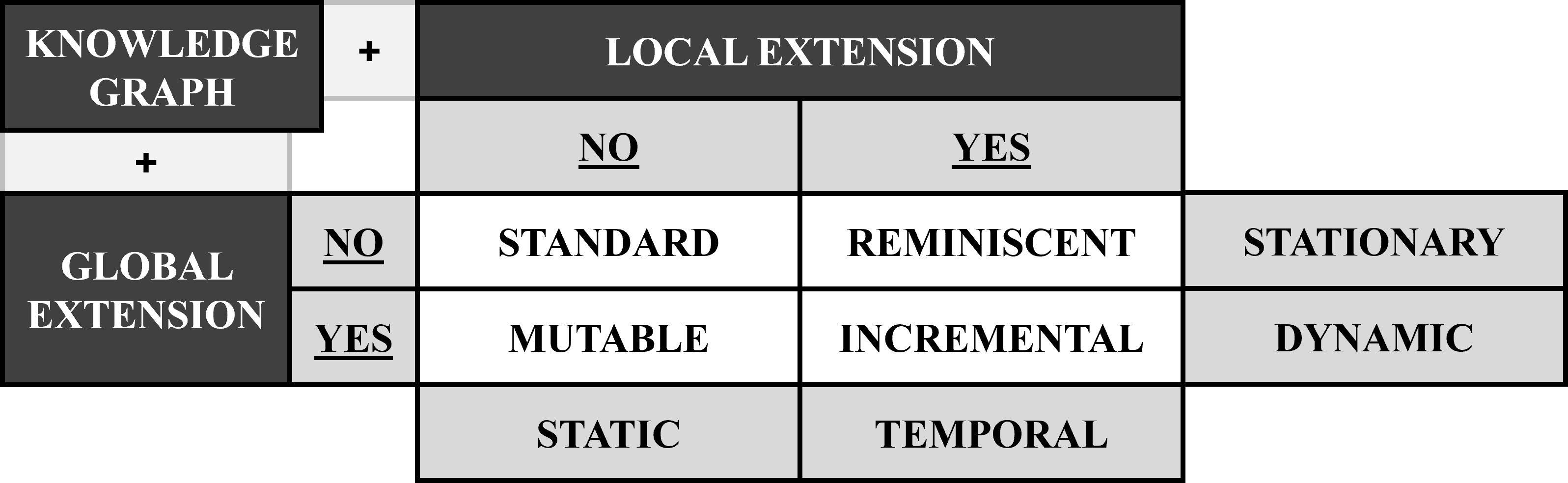}
\caption{Overview of the time-aware knowledge graph extensions regarding possible combinations of local extensions (edge timestamps) and global extensions (consideration of multiple consecutive versions of a graph).}
\label{fig:overview}
\end{figure}
\section{Time-Aware Knowledge Graph Extensions}
In this section, a generalized framework for extending KGs with time-awareness is introduced, which intends to cover existing and future work in this field. As indicated in Section~\ref{relatedwork}, we are concerned with local extensions, i.e., enriching edges with timestamps, as well as global extensions, i.e., considering the evolution of a graph.
The overview in Figure~\ref{fig:overview} shows the different types of time-aware KGs introduced in this paper. According to the most common terms in the literature, we refer to locally extended KGs as temporal and globally extended KGs as dynamic. If the respective extension is not considered, then the KG is referred to as static and/or stationary. For example, a standard KG is static because all edges in the graph must be interpreted as static triples, and stationary because it models domain knowledge with respect to a fixed point in time.

We investigate different types of KG extensions. A reminiscent KG models the domain knowledge for a fixed point in time, but is provided with memory in the form of additional edge timestamps. Regarding a fixed timestamp, mutable KGs do not contain this memory, but can be observed over the time period $\mathcal{T}$. As a combination, incremental KGs are equipped with additional edge metadata and are observable with respect to $\mathcal{T}$.

\subsection{Stationary Knowledge Graphs}
\label{stationary}
Standard KGs represent special cases of static and stationary KGs as they are stationary instantiations of static ontologies $\mathcal{O} = (\mathcal{C}, \mathcal{L}, \mathcal{R}, \rho)$. 
Such ontologies include concepts $\mathcal{C}$, i.e. entity types, attributive literals $\mathcal{L}$, as well as attributive and contextual relations in $\mathcal{R}$.
In addition, $\rho$ denotes the instantiation rules of $\mathcal{O}$ which assess whether triples are valid or not, based on the triples themselves and the topology of the graph. For example, a member of the $EU$ must necessarily be a country.
Given such an ontology $\mathcal{O}$, a triplestore including a set of entites $V$ and a set of edges $E$ with entries $(h,r,t)$ is called a standard knowledge graph $\mathcal{G} = (V,E)$ if $h \in V$, $r \in \mathcal{R}$, and $t \in V$ or $t \in \mathcal{L}$ holds such that the validity of triples can be assessed using the rules in $\rho$. In some works, blank nodes are considered as head or tail nodes, but we omit this here without loss of generality, since they can always be added by including the concept of a blank node in $\mathcal{C}$. To extend the facts in $\mathcal{G}$ with additional timestamps, an adaptation of static ontologies is required.
\begin{definition}[Temporal Ontology]
Given a time set $\mathcal{T}$ as well as a standard ontology $\mathcal{O} = (\mathcal{C}, \mathcal{L}, \mathcal{R}, \rho)$, a temporal ontology is defined as $\mathcal{O}^+ = (\mathcal{C}, \mathcal{L}, \mathcal{R}, \mathcal{T}, \rho^+)$ such that triples $(h,r,t)$ are replaced by quintuples $(h,r,t,\tau_{start}, \tau_{end})$ with additional timestamps $\tau_{start}, \tau_{end} \in \overline{\mathcal{T}}$, defining the start and end of validity of an edge. Further, the instantiation rules $\rho$ are extended by time-related rules which determine the validity of quintuples.
\end{definition}
Regarding the additional time-related rules in $\rho^+$, there are some obvious rules, such as the one that $\tau_{start} \leq \tau_{end}$ should always hold. To ensure this rule also for quintuples for which no additional time-related information is available, we assume the previously introduced closure $\overline{\mathcal{T}}$, so that $\tau_{start} = -\infty$ or $\tau_{end} = \infty$ may be used if necessary. By means of $\mathcal{T} = \emptyset$ and $\overline{\mathcal{T}} = \{-\infty,\infty\}$, one also recognizes that static ontologies represent special cases of temporal ontologies. Therefore, with respect to temporal ontologies, we always assume $\mathcal{T} \neq \emptyset$.
Accordingly, a real-world instantiation of a temporal ontology for a fixed timestamp is called a reminiscent knowledge graph which is defined as follows.
\begin{definition}[Reminiscent Knowledge Graph]
Let $\mathcal{O}^+ = (\mathcal{C}, \mathcal{L}, \mathcal{R}, \mathcal{T}, \rho^+)$ be a temporal ontology. Then a quintuple store including entites $V$ and a set of edges $E^+$ with entries $(h,r,t, \tau_{start}, \tau_{end})$ is a reminiscent knowledge graph $\mathcal{G}^+ = (V,E^+)$ if $h \in V$, $r \in \mathcal{R}$, $t \in V$ or $h \in \mathcal{L}$, and $\tau_{start}, \tau_{end} \in \overline{\mathcal{T}}$ holds and there are no violations of the rules in $\rho^+$.
\end{definition}

Similar graph implementations as special cases of the above definition are already used in existing works, for example to optimize KG embeddings with temporal aspects or to integrate past information in a KG and successive applications~\cite{temp_emb,temp_completion}. Since often only single timestamps $\tau_{start}$ are considered, we introduce the notion of \textit{semi-temporality} which is present if $\tau_{end} = \infty$ holds for all quintuples. Accordingly, we introduce \textit{semi-reminiscent} KGs as special cases of reminiscent KGs. However, in this case, the deactivation of an edge inevitably leads to its deletion, i.e., only active edges are present.

\subsection{Dynamic Knowledge Graphs}
Unlike standard and reminiscent knowledge graphs as stationary domain representations, many KG applications are meant to go beyond the original encoding of semistructured data for a fixed point in time. For example, KG embeddings and graph neural networks are supposed to be adaptive such that they can be efficiently reused after the topology of the KG is updated~\cite{puTransE,DKGE}. Therefore, it is necessary to consider KG representations that are dynamic with respect to a set of timestamps $\mathcal{T}$, which justifies the following definition of a dynamic KG as a mapping from $\mathcal{T}$ to an appropriate set of KGs.

\begin{definition}[Dynamic Knowledge Graph]
We assume a set of timestamps $\mathcal{T}$, a set of either static or temporal ontologies $\{\mathcal{O}_\tau~:~\tau \in \mathcal{T} \}$ with corresponding sets $\mathbb{G}_\tau$ of all stationary KGs according to $\mathcal{O}_\tau$ and we define $\mathbb{G} := \bigcup_{\tau \in \mathcal{T}} \mathbb{G}_\tau$. Then, a dynamic knowledge graph is defined as a mapping $\Gamma: \mathcal{T} \rightarrow \mathbb{G}$ such that $\Gamma (\tau ) \in \mathbb{G}_\tau$ holds for all $\tau \in \mathcal{T}$.
\label{def:dynamic}
\end{definition}
Apparently, Definition~\ref{def:dynamic} does not specify whether static or temporal ontologies are considered. Figure~\ref{fig:overview} suggests that this may result in different types of time-aware KG extensions, which we discuss in the following. First, we assume static ontologies.
\begin{definition}[Mutable Knowledge Graph]
A dynamic knowledge graph $\Gamma$ is called a mutable knowledge graph if the underlying ontologies $\{\mathcal{O}_\tau:~\tau \in \mathcal{T} \}$ are static ontologies and $\Gamma(\tau)$ yields a standard knowledge graph for all $\tau \in \mathcal{T}$.
\end{definition}
A mutable KG thus offers the possibility to infer when triples were added to the graph, i.e., when the inherent information became accessible. However, the stationary images $\Gamma (\tau)$ for $\tau \in \mathcal{T}$ are static and therefore do not contain information about the period of validity of an edge. Thus, regarding time-awareness in general, one has to distinguish between the validity and the accessibility of facts to the knowledge graph.
\begin{definition}[Incremental Knowledge Graph]
A dynamic KG $\Gamma$ is called an incremental knowledge graph and is denoted as $\Gamma^+$ if the underlying ontologies $\{\mathcal{O}_\tau:~\tau \in \mathcal{T} \}$ are temporal ontologies and $\Gamma^+(\tau)$ yields a reminiscent KG for all $\tau \in \mathcal{T}$.
\end{definition}
An incremental KG thus offers the possibility to trace the accessibility of facts as well as their validity. Since deactivated edges are kept in the graph, for timestamps $\tau, \tau' \in \mathcal{T}$ with $\tau < \tau'$, $\Gamma^+ (\tau')$ contains at least as many facts as $\Gamma (\tau)$. Therefore, according to the notion of semi-temporality from Section~\ref{stationary}, we also allow \textit{semi-incremental} KGs as special cases of incremental KGs whose stationary images $\Gamma^+(\tau)$ are semi-reminiscent, i.e., they only contain active facts. However, assuming the prior existence of an edge, the time of its deletion is reconstructable by inspecting the previous versions of the graph.

\subsection{Application Example}
As indicated in Section~\ref{relatedwork}, standard KGs are not sufficient to encode temporally extended facts like the membership of the UK in the EU from $1973$ to $2020$. Therefore, reminiscent KGs with additional timestamps $\tau_{start}, \tau_{end} \in \overline{\mathcal{T}}$ are considered so that the fact is encodable as $(UK, member, EU,1973,2020)$. In addition, if only active, i.e. currently valid edges are to be contained in the graph, a semi-reminiscent graph can be implemented.

However, stationary KGs do not not trace the time-related evolution of the information encoded within a graph, since they only contain the knowledge available for a fixed timestamp. Therefore, dynamic KGs are introduced in this paper to account for such evolutions. In Table~\ref{tab:tab}, the different dynamic extension types are exemplified. Thus, in 2012, the information that the UK is a member of the EU since $1973$ is added to each KG. However, the year $1973$ is only explicitly encodable in the (semi-)incremental KGs such that this additional information is still available in $2020$. Due to its staticness, this is not the case for the mutable KG. After the Brexit in $2020$, the information about the previous membership is removed from the mutable and the semi-incremental KG, since they only contain active edges. In the incremental KG, on the other hand, the fact is updated by means of the new timestamp $\tau_{end}=2020$, making it available for future timestamps $\tau > 2020$ as well. However, this also leads to a continual expansion of the graph.
\subsection{Time-Awareness in Existing Knowledge Graphs and Applications}
Many existing and established knowledge graphs such as DBpedia, YAGO, and Wikidata already satisfy certain requirements for time-awareness. Indeed, KGs are mostly based on dynamic domains such that a versioning or the implementation and utilization of a SPARQL update endpoint results in a dynamic KG. However, these dynamics are mostly not explicitly considered, but rather the stationary images of the graphs. In particular, successive applications, such as embeddings and graph neural networks, are developed or trained for stationary images without taking the underlying dynamics into account.
\begin{table}[b!]
\caption{Application of dynamic, i.e., \textbf{m}utable (\textbf{m}), \textbf{s}emi-\textbf{i}ncremental (\textbf{s-i}) and \textbf{i}ncremental (\textbf{i}) KG extensions.}
\label{tab:tab}
\begin{tabular}{|c|c|c|c|c|}
\hline
~ & 2012 & ... & 2020 & 2021\\
\hline
\textbf{m} & $(UK, member, EU)$ & & $(UK, member, EU)$ & - \\
\textbf{s-i} & $(UK, member, EU,1973,\infty)$ & & $(UK, member, EU,1973,\infty)$ & - \\
\textbf{i} & $(UK, member, EU,1973,\infty)$  & & $(UK, member, EU,1973,\infty)$ & $(UK, member, EU,1973,2020)$ \\
\hline
\end{tabular}
\end{table}
\section{Summary}
This work contributes to the Semantic Web community by providing a generalized framework for time-aware knowledge graph extensions. Current research shows that further progress is needed to establish knowledge graphs in non-stationary and non-static domains. In this context, some promising approaches already exist that extend or adapt methods for standard knowledge graphs to make them usable for dynamic or temporal KGs as well. However, so far, these approaches do not share a common vocabulary to compare their respective results. The definitions introduced in this work provide this kind of vocabulary to facilitate the comparison and integration of existing, but also future works and thus can serve as an accelerator for the desired progress of knowledge graph implementations and applications in time-aware, i.e., dynamic and temporal domains.
\\\\
\textbf{Acknowledgements.} This work is part of the TEAMING.AI project which receives funding in the European Commission's Horizon 2020 Research Programme under Grant Agreement Number 957402 (www.teamingai-project.eu).
\bibliographystyle{vancouver}
\bibliography{bibliography}

\begin{thebibliography}{10}

\bibitem{QA}
Diefenbach D, Gim{\'e}nez-Garc{\'i}a J, et~al.
\newblock QAnswer KG: Designing a Portable Question Answering System over RDF
  Data.
\newblock In: The Semantic Web: ESWC 2020; 2020. p. 429-45.

\bibitem{recommendation}
Palumbo E, Rizzo G, et~al.
\newblock Knowledge Graph Embeddings with node2vec for Item Recommendation.
\newblock In: The Semantic Web: ESWC 2018 Satellite Events; 2018. p. 117-20.

\bibitem{kg_for_ind4.0}
Bader SR, Grangel-Gonzalez I, et~al.
\newblock A Knowledge Graph for Industry 4.0.
\newblock In: The Semantic Web: ESWC 2020; 2020. p. 465-80.

\bibitem{biomedicine}
Vidal ME, Endris KM, et~al.
\newblock Semantic Data Integration of Big Biomedical Data for Supporting
  Personalised Medicine.
\newblock In: Current Trends in Semantic Web Technologies: Theory and Practice.
  Springer International Publishing; 2019. p. 25-56.

\bibitem{rdf-star}
Hartig O.
\newblock Foundations of RDF{\(\star\)} and SPARQL{\(\star\)} (An Alternative
  Approach to Statement-Level Metadata in {RDF)}.
\newblock In: Proceedings of the 11th Alberto Mendelzon International Workshop
  on Foundations of Data Management and the Web. 2017.

\bibitem{context-aware}
Liu Y, Hua W, et~al.
\newblock Context-Aware Temporal Knowledge Graph Embedding.
\newblock In: Web Information Systems Engineering -- WISE 2019; 2019. p.
  583-98.

\bibitem{triedf}
Pelgrin O, Gal{\'a}rraga L, Hose K.
\newblock TrieDF: Efficient In-memory Indexing for Metadata-augmented RDF.
\newblock In: Proceedings of the 7th Workshop on Managing the Evolution and
  Preservation of the Data Web (MEPDaW) co-located with ISWC 2021; 2021. p.
  1-10.

\bibitem{towards_kg}
Ehrlinger L, W{\"o}{\ss} W.
\newblock Towards a Definition of Knowledge Graphs.
\newblock In: Joint Proceedings of the Posters and Demos Track of SEMANTiCS2016
  and SuCCESS'16. 2021.

\bibitem{kg_survey}
Hogan A, Blomqvist E, et~al.
\newblock Knowledge Graphs.
\newblock ACM Comput Surv. 2021;54(4).

\bibitem{temp_emb}
Xu C, Nayyeri M, et~al.
\newblock Temporal Knowledge Graph Completion Based on Time Series Gaussian
  Embedding.
\newblock In: The Semantic Web -- ISWC 2020; 2020. p. 654-71.

\bibitem{temp_completion_1}
Garc{\'\i}a-Dur{\'a}n A, Duman{\v{c}}i{\'c} S, Niepert M.
\newblock Learning Sequence Encoders for Temporal Knowledge Graph Completion.
\newblock In: Proceedings of the EMNLP 2018; 2018. p. 4816-21.

\bibitem{temp_completion}
Xu C, Chen YY, et~al.
\newblock Temporal Knowledge Graph Completion using a Linear Temporal
  Regularizer and Multivector Embeddings.
\newblock In: Proceedings of the 2021 Conference of the North American Chapter
  of the Association for Computational Linguistics: Human Language
  Technologies; 2021. p. 2569-78.

\bibitem{puTransE}
Tay Y, Luu A, et~al.
\newblock Non-Parametric Estimation of Multiple Embeddings for Link Prediction
  on Dynamic Knowledge Graphs.
\newblock Proceedings of the AAAI Conference on Artificial Intelligence.
  2017;31(1).

\bibitem{DKGE}
Wu T, Khan A, et~al.
\newblock Efficiently Embedding Dynamic Knowledge Graphs.
\newblock ArXiv. 2019;~abs/1910.06708.

\bibitem{dyn_compl}
Xie W, Wang S, et~al.
\newblock Dynamic Knowledge Graph Completion with Jointly Structural and
  Textual Dependency.
\newblock In: Algorithms and Architectures for Parallel Processing; 2020. p.
  432–448.

\end{thebibliography}
\end{document}